\documentclass[10pt,letterpaper]{article}
\usepackage[top=0.85in,left=2.75in,footskip=0.75in,marginparwidth=2in]{geometry}

\usepackage[utf8]{inputenc}

\usepackage{cite}

\usepackage{algorithm2e}

\usepackage{nameref,hyperref}

\usepackage[right]{lineno}

\usepackage{microtype}
\DisableLigatures[f]{encoding = *, family = * }

\raggedright
\usepackage{changepage}

\usepackage[aboveskip=1pt,labelfont=bf,labelsep=period,singlelinecheck=off]{caption}

\makeatletter
\renewcommand{\@biblabel}[1]{\quad#1.}
\makeatother

\usepackage{lastpage,fancyhdr,graphicx}
\usepackage{epstopdf}
\fancyheadoffset[L]{2.25in}
\fancyfootoffset[L]{2.25in}

\usepackage{color}

\definecolor{Gray}{gray}{.25}

\usepackage{graphicx}
\usepackage{multirow}
\usepackage{amsmath}
\usepackage{amssymb}

\usepackage{sidecap}

\usepackage{wrapfig}
\usepackage[pscoord]{eso-pic}
\usepackage[fulladjust]{marginnote}
\reversemarginpar

\usepackage{tikz}
\usepackage{textcomp}
\usepackage{hyperref}
\usepackage{lipsum}

\newcommand\copyrighttext{%
  \footnotesize \textcopyright 2012 IEEE. Personal use of this material is permitted.
  Permission from IEEE must be obtained for all other uses, in any current or future
  media, including reprinting/republishing this material for advertising or promotional
  purposes, creating new collective works, for resale or redistribution to servers or
  lists, or reuse of any copyrighted component of this work in other works.}
\newcommand\copyrightnotice{%
\begin{tikzpicture}[remember picture,overlay]
\node[anchor=south,yshift=10pt] at (current page.south) {\fbox{\parbox{\dimexpr\textwidth-\fboxsep-\fboxrule\relax}{\copyrighttext}}};
\end{tikzpicture}%
}

\begin{document}
\vspace*{0.35in}


\begin{flushleft}
{\Large
\textbf\newline{Ischemic Stroke Identification Based on EEG and EOG
using 1D Convolutional Neural Network and Batch Normalization}
}
\newline
\\
Endang Purnama Giri\textsuperscript{1,2},
Mohamad Ivan Fanany\textsuperscript{1},
Aniati Murni Arymurthy\textsuperscript{1},
\\
\bigskip
\bf{1} Machine Learning and Computer Vision Laboratory, \\Faculty of Computer Science, Universitas Indonesia\\
\bf{2} Computer Sciences Department, Faculty of Mathematics \\and Natural Sciences,
Bogor Agricultural University\\
\bigskip
* epgthebest@gmail.com

\end{flushleft}

\copyrightnotice

\providecommand{\keywords}[1]{\textbf{\textit{Keywords---}} #1}

\section*{Abstract}
In 2015, stroke was the number one cause of death in Indonesia. The majority type of stroke is ischemic. The standard tool for diagnosing stroke is CT-Scan. For developing countries like Indonesia, the availability of CT-Scan is very limited and still relatively expensive. Because of the availability, another device that potential to diagnose stroke in Indonesia is EEG. Ischemic stroke occurs because of obstruction that can make the cerebral blood flow (CBF) on a person with stroke has become lower than CBF on a normal person (control) so that the EEG signal have a deceleration. On this study, we perform the ability of 1D Convolutional Neural Network (1DCNN) to construct classification model that can distinguish the EEG and EOG stroke data from EEG and EOG control data. To accelerate training process our model we use Batch Normalization. Involving 62 person data object and from leave one out the scenario with five times repetition of measurement we obtain the average of accuracy 0.86 (F-Score 0.861) only at 200 epoch. This result is better than all over shallow and popular classifiers as the comparator (the best result of accuracy 0.69 and F-Score 0.72 ). The feature used in our study were only 24 ‘handcrafted’ feature with simple feature extraction process.
\bigskip

\noindent\keywords{Sleep stage, Conditional Neural Fields, Deep Belief Networks, Fuzzy C-Means Clustering, Classification}


\section*{Introduction}
\label{Introduction}

Stroke is classified into two types: ischemic and hemorrhagic. For a specific ischemic stroke occurs when a blockage (obstruction) of the small blood vessels around the brain (Fig. 1). Furthermore, ischemic is the majority of stroke events. Ischemic stroke is 87\% of all stroke cases in the United States in 2015, as reported by American Stroke Association 2015 \cite{ASA}. Based on the results of research published by the Agency for Health Research and Development (Balitbangkes) in January 2015, stroke was the number one of disease that leading cause of death in Indonesia \cite{tempo}. The prevalence of stroke in Indonesia keeps on the increase. On Research Publication of Basic Health (Riskesda) 2013 the prevalence had reached 12.1\%. The province in Indonesia with the highest prevalence of stroke and its symptoms in the year 2013 was South Sulawesi (17.9\%) \cite{balitbangkes}. In general, large urban areas recorded a higher prevalence of stroke than the rural area. Some people believe that the higher prevalence happened because of the pressure of life in big cities are greater than the pressure of life in the countryside. However, as the concern is in the big city which has medical support equipment that widely available so that the incidence of stroke is recorded better than the data collection in rural areas that have inadequate health facilities.
\begin{figure}
\includegraphics[scale = 0.4]{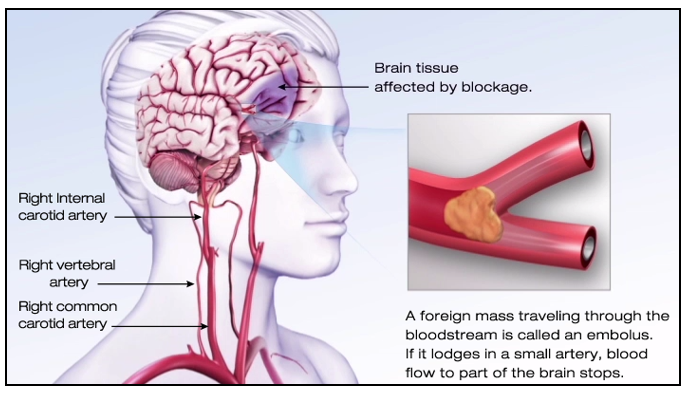}
\caption{Ischemic stroke illustration (American Stroke Association 2015).}
\label{fig:isc}
\end{figure}
\par 
The tool that can be used to diagnose stroke accurately is a Computerized Tomography Scanner (CT-Scan). Although stated at safe levels, the use of CT-Scan procedure still quite risky, because the patient is exposed by X-rays. Especially for developing countries like Indonesia the availability of CT-Scan is still very limited, only available in large hospitals. Besides of that, the cost of CT-Scan use is still relatively expensive. Another potential tool to be used as an alternative or support the CT-Scan is Electroencephalography (EEG). Not only in addition to minimizing the risk but also the availability of EEG in Indonesia is more than the availability of CT scan. EEG have been already available in a hospital with grade C. The hospital with grade C is a hospital with limited health care subspecialists which is available in each district in Indonesia.

\par 
Based on the conditions described above, it can be identified an issue related to the handling of stroke in rural areas in Indonesia with inadequate health facilities. Because of in Indonesia, the presence of EEG relatively more available than CT scan so an alternative solution that can be taken is via addition the usefulness of EEG for detecting ischemic stroke. Unfortunately, until now there are no accurate algorithms as the standard that can be used to detect and identify the ischemic stroke events through the EEG signals. When a robust algorithm for detecting the ischemic stroke based on EEG signal can be formulated, then it makes the EEG be a better choice to use than a CT scan in Indonesia. Referring to the problems in this article we present new techniques for recognizing the ischemic stroke from the EEG signal using deep learning approach. 

\section{Previous Research}

The diagnosis of ischemic stroke by EEG rhythms often associated with the function of blood circulation around the brain as a carrier of oxygen to the brain. Blood circulation in the brains of normal people has a level of Cerebral Blood Flow (CBF): 50-70 mL.100 g-1.min-1. For ischemic stroke patients, they have indications that the CBF value only in the range of between 25-30 g-1.min mL.100-1 \cite{Jordan}. Through the observation, the condition of impairment CBF is accompanied by changes in the pattern of EEG signals. The declining value of the CBF will be accompanied by the decelerating of the EEG signals. EEG signal patterns undergo changes ranging from the loss of fast beta waves (fast frequency like beta and gamma), followed by changes theta rhythm become dominant, until getting slow down and become to be delta waves.

From some previous studies revealed that it is possible to diagnose ischemic stroke use EEG. Murri et al. in \cite{muri} have proven that Quantitative EEG (qEEG) has expanded the functionality of a conventional EEG. qEEG is the result of numerical analysis of the EEG signal spectrum. In that study also analyzes the correlation between interpretation value of qEEG compare to CT scan diagnose for patients with ischemic stroke. As a conclusion qEEG afford: (i) to support the localization of parts of nerve on clinical examination, ii) qEEG provide prognostic index of functional disorders of the brain and iii) qEEG can construct a 'basic electrophysiology' to determine further action on the patient with ischemic. However, the analysis process using qEEG rated requires time and need some stages, so it is relatively difficult to implement. Jordan in \cite{Jordan} also suggests the potentiality use of EEG combined with cutting-edge signal processing techniques for early detection of Acute Ischemic Stroke (AIS) in the future. His statement was supported by the evidence that the analysis of the EEG signal can add to the value of early diagnosis, prediction results, the selection of medication treatment for patients, clinical management, and detection of seizures in AIS. Foreman and Claassen \cite{fore} found preliminary evidence that quantitative EEG (qEEG) is quite sensitive for the detection of ischemic stroke to avoid permanent nerve damage. Referring to it, on this article, we perform the ability of EEG combine with deep learning techniques to recognize the signal structure of an ischemic stroke in EEG signal. From this study we hope that the use of EEG can be expanded and in some circumstances can be used for ischemic stroke diagnose.

\begin{equation}
BSI(t)=\frac{1}{M}\sum_{j=1}^{M}\left \| \sum_{i-1}^{N}\frac{R_{ij}(t)-L_{ij}(t)}{R_{ij}(t)-L_{ij}(t)} \right \|
\label{eq:bsi}
\end{equation}

Another study that was conducted by Putten et al. \cite{Putten} have proposed a measure quantification of EEG signal as Brain Symmetry Index (BSI). BSI value describes the size of the symmetry of condition parts of the left brain and the right brain, given by (\ref{eq:bsi}). The symmetry of the value represented as the average value of the absolute value of the average difference between the intensity of hemispheric power at frequencies with a range of 1-25Hz. As an estimator to calculate the value of the power of hemispheric, it can use  Fast Fourier Transform (FFT). In the research done by Putten et al., they determined the threshold value of BSI for healthy people is below 0.042$\pm$0.005. A significant result of their study is they can quantitatively conclude that the value of BSI positively correlated to the value of NIHSS. The correlation coefficient between BSI and NIHSS is high enough which is $\rho \approx$ 0.86 (P$<$0.01). Conclusively the relationship between BSI and NIHSS can be formulated as BSI = 0.0077 NIHSS + 0.044. National Institutes of Health Stroke Scale (NIHSS) is a standard measurement (a scoring system) that is commonly used by health care providers when measuring the impact of the functional impairment of the body as an impact of a stroke \cite{hage}. In 2015 \cite{sastra} another study was conducted by Sastra et al. using Welch analysis for detecting the decrement of frequency and BSI (brain symmetry index) for the left and right hemisphere of the brain and compared with the CT-scan to get the abnormalities of the signals. There were some irregularities of the EEG signals in comparisons with CT-scan. All patients were examined stroke based on CT-scan. However, there were examined as normal, epileptic form, and stroke based on raw EEG readings by the neurologist. All of the BSI calculations were above the healthy subjects (0.042$\pm$0.005), which indicated that all subject were Acute Ischemic Stroke. While for some readings according to the conventional EEG indicated that 20\% of the subjects were normal or within normal conditions. These were consistent with the ratios of power densities that all subjects were abnormal. These results were consistence with the CT Scans, but they should be extended to more subjects to get better conclusions.

\section{Data, Features, Classifiers, and Batch Normalization}
This section describes the data, features, and several classifiers to be compared in our studies.
\subsection{Data}
We provide and use primary EEG data that were recorded by our collaboration with Pusat Otak Nasional (PON) or National Brain Center Hospital, Jakarta. The data involve 32 patients with stroke and 30 normal persons as data control. For this study, we use two class labels that are ’Stroke’ and ’Normal.' For each data object was recorded signals from 33 channels. For data acquisition, we use two EEG tools. One machine with brand Xltek and the other one is Biologic. The sampling rate of EEG signal for Xltek is 512Hz, and for Biologic we use 512Hz for some data and 256Hz for the other. The duration time for each EEG measurement were 15 minutes in total. On this study, we only use 2 channels EEG (C3 and OZ), and 2 electrooculography channels (left EOG and right EOG).     
\subsection{Features}
This study uses  24 ‘handcrafted’ features. All of these features are the features that a part of features was used in \cite{Lankvist}. Different with \cite{Lankvist} this study not involve band frequency of delta. In detail the features composed by: 
\begin{enumerate}
    \item \textbf{Relative power for four band frequencies:} delta (0.5$-$4 Hz), theta (4$-$8 Hz), alpha (8$-$13 Hz), and beta (13$-$20 Hz). This relative power feature amounted to 12 features that consist of: (f01) C3 delta, (f02) C3 theta, (f03) C3 alpha, (f04) C3 beta, (f05) EOG delta, (f06) EOG theta, (f07) EOG alpha, (f08) EOG beta, (f09) OZ delta, (f10) OZ theta, (f11) OZ alpha, and (f12) OZ beta.

    \item \textbf{Standard deviation EOG }(f13): calculate as varian between each value on EOG signals.
    
    \item \textbf{EOG Correlation:} (f14) Correlation of EOG (EOGcorr)
    
    \item \textbf{Value of Kurtosis:} we use three values of kurtosis (f15) C3 kurtosis, (f16) EOG kurtosis, and (F17) OZ kurtosis.
    
    \item \textbf{Entropy:} In this study, we use three values of entropy (f18) C3 entropy, (f19) EOG entropy, and (f20) OZ entropy.
    
    \item \textbf{Mean of spectral:} In this study, we use three values of the spectral mean (f25) EEG spectral mean, (f21) EOG spectral mean, and (f22)  OZ spectral mean.
    
    \item \textbf{Fractal Exponent:} for the last one is (f24), C3 fractal exponent that is calculated as the negative slope of the linear fit of spectral density in the double logarithmic graph.
\end{enumerate}

\subsection{Classifiers}
In this study, we compare 1D Convolutional Neural Network (1DCNN) to several shallow and popular classifiers. This section shortly describes 1DCNN and the list of shallow classifiers as a comparator.
\subsubsection{1DCNN} 
\begin{figure}
\includegraphics[scale = 0.35]{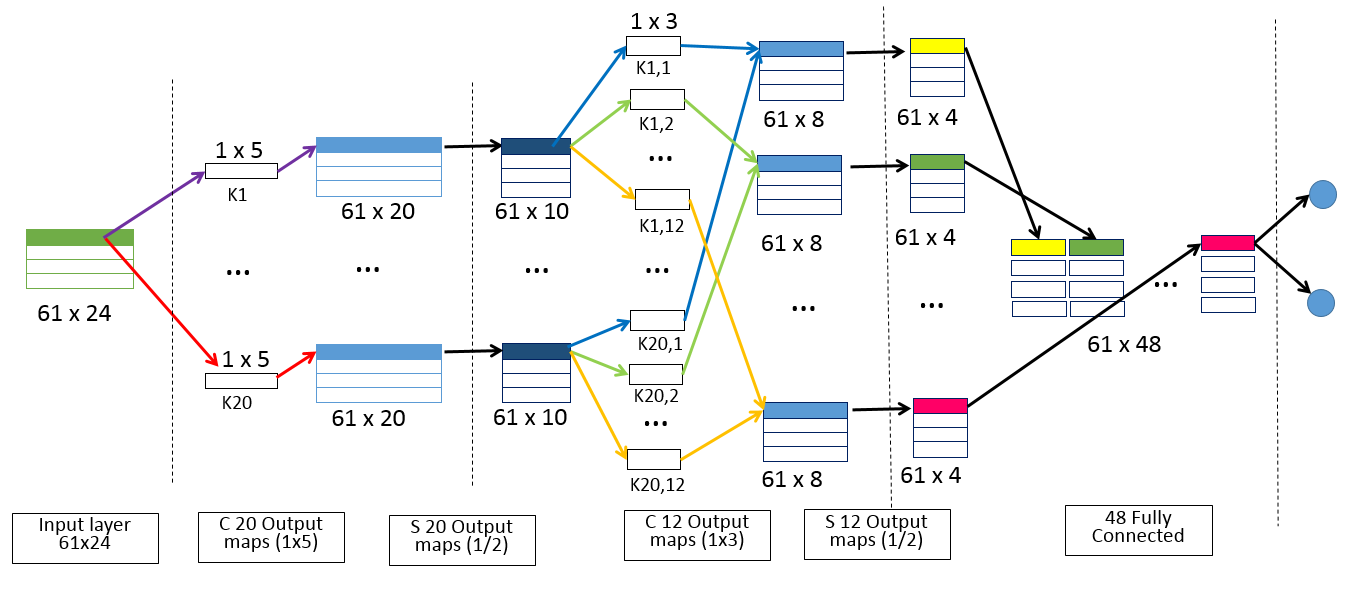}
\caption{Feed Forward 1DCNN 20C-20S-12C-12S with kernel size 1x5 for C on layer 1 and 1x3 for C on layer 4.}
\label{fig:ff}
\end{figure}

\begin{figure}
\includegraphics[scale = 0.35]{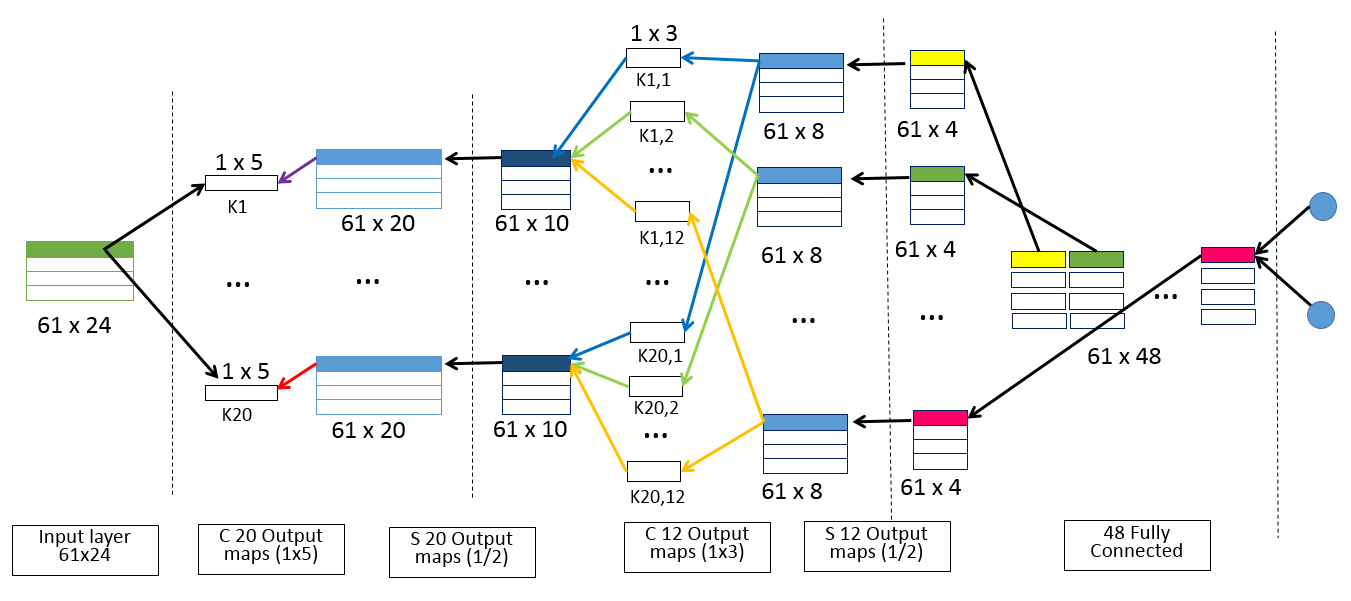}
\caption{Back Propagation 1DCNN 20C-20S-12C-12S with kernel size 1x5 for C on layer 1 and 1x3 for C on layer 4.}
\label{fig:bp}
\end{figure}

is one of an example of a classifier that categorized as a discriminative architecture of deep learning algorithm. Two main procedures of CNN is convolutional function and pooling operation \cite{zheng}. In our study, 1D convolutional operation is useful to extract the important local feature in between neighbouring element value of feature vector. On the other hand 1D pooling operation view globally and generalize the value of information for each important local feature. With backpropagation approach, the weight of 1D kernel value of convolution will be automatic update and determine. The architecture 1DCNN that was used in our study given by Fig. \ref{fig:ff} and Fig. \ref{fig:bp}. On that figure, the architecture of our 1DCNN are:

\begin{itemize}
    \item \textbf{1st layer\::} 20  convolution maps with kernel size 1x5
    \item \textbf{2nd layer:} 20  subsampling maps with scaling factor 1/2
    \item \textbf{3rd layer:} 12  convolution maps with kernel size 1x3
    \item \textbf{4th layer:} 12  subsampling maps with scaling factor 1/2
    \item \textbf{5th layer:} fully connected layer with number of neurons 48
\end{itemize}
In this study, we also add batch normalization procedure before activation layer on the last layer (between the 4th layer and 5th layer). Another technique was used to obtain faster construction model process than usual technique, in this study, we also apply early stopping method when training process was running. The stop criteria in our study are when the model can predict the data testing correctly. On this procedure, we evaluate each resulting model from each epoch. When the model can predict correctly, the training process stops on that epoch. For the experiment, we use 1DCNN that implementation with Keras and Theano python library \cite{chollet2015keras}.

\subsubsection{Popular Shallow Classifier}
We use several shallow (architecture) classifiers to check whether or not the problem of ischemic stroke EEG classification is a hard problem. Shallow classifiers that were utilized in this study are Naive Bayes (NB), kNN (Nearest Neighbour), Random Forest (RF), Classification Tree (CT), Neural Network (NN), and Logistic Regression (LogReg). Via experiment the performance of each classifier is measured and then compared to the performance of 1DCNN. Some of this shallow classifiers were used in \cite{Giri} and perform good enough ability on a lot of classification task. On the experiment, we use Orange for data mining to perform all of these comparator classifiers \cite{orange}.
Parameters for each classifier are set for the best result by trying some setting combinations. For details the parameter setting for all comparative classifiers are:
\par
\begin{itemize}
    \item \textbf{Naive Bayes\::} Probability estimation use prior relative frequency with 100 LOESS sample points and size of LOESS  window 0.5
    \item \textbf{kNN:} number of cluster 2, minimum number of neighbours 5, and euclidian distance metrice.
    \item \textbf{Random Forest:} Number of tree in forest is 10, with stop splitting node when the instance is fewer than 5.
    \item \textbf{Classification Tree:} with attribute selection criteria for split is information gain.
    \item \textbf{Neural Network:} use 24 input neurons, 200 hidden neurons, and 2 output neurons, regularization factor 0.1 and number of epoch 200
    \item \textbf{Logistic Regression:} use regularization technique squared weights with training error cost 1.00.
\end{itemize}

\begin{equation}
input : B= \big\{ x_{1...m} \big\}; 
\label{eq:v}
\end{equation}

$\gamma$ and  $\beta$ as to be train parameter

\begin{equation}
 output: \big\{y_{i}=BN_{ \gamma , \beta }(x_{i})\big\}   
\label{eq:v}
\end{equation}

\begin{equation}
\mu _{B} \leftarrow  \frac{1}{m}  \sum_{i=1}^{m} x_{i}  
\label{eq:v}
\end{equation}

\begin{equation}
\sigma_{B}^{2} \leftarrow  \frac{1}{m}  \sum_{i=1}^{m} \left (  x_{i}-\mu_B\right )^{2} 
\end{equation}

\begin{equation}
 \widehat{x_{i}}  \leftarrow  \frac{x_{i}-\mu_{B}}{ \sqrt{\sigma_{B}^{2}+ \epsilon} }
\end{equation}

\begin{equation}
  y_{i}\leftarrow \gamma \widehat{x_{i}}+\beta \equiv
  BN_{ \gamma , \beta }(x_{i})
\end{equation}

\subsection{Batch Normalization}
Many layers of deep neural architecture make the training process difficult. Deep architecture needs small value learning rate and much more time to training. It is hard to train models with saturating nonlinearities on deep neural architecture. This condition is called as internal covariate shift problems. One of technique to solve this issue is by normalization input layer. As mention on \cite{BN}, the main goal of Batch Normalization is to achieve a stable distribution of activation values throughout training. Batch normalization applies normalization technique for each training mini batch. 

\par For general algorithm steps of batch normalization given by (2) to (7). For each equation represent each aspect of the normalization process.Equation (4) is mini batch mean, Equation (5) is mini batch variance, Equation (6) is normalization task, and Equation (7) is scale and shift process as learning process to train $\gamma$ and  $\beta$. Batch normalization procedure can be applied to each activation layer. From several previous research and studies, Batch Normalization has been proven can address internal covariate shift problem, so that we can reduce the number of iteration towards convergence, hence accelerate the training process.

\section{Methodology}
\begin{figure}
\includegraphics[scale = 0.6]{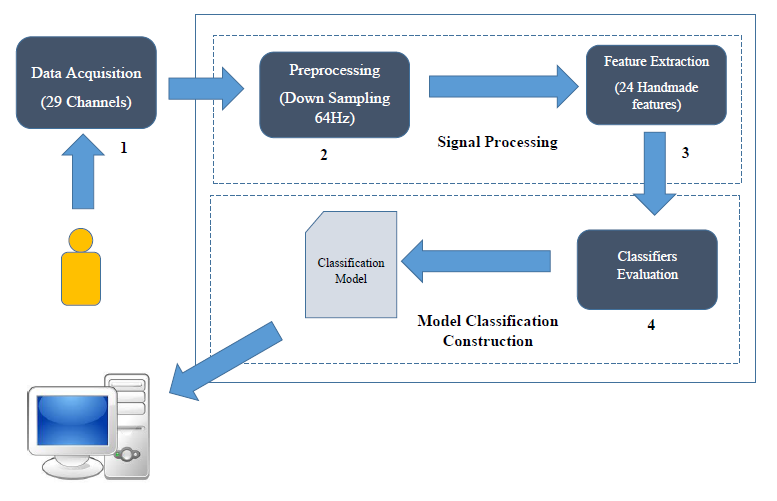}
\caption{Methodology.}
\label{fig:method}
\end{figure}

This section will describe four main steps in our methodology (Fig. \ref{fig:method}).

\subsection{Data Acquisition}

\begin{figure}
\includegraphics[scale = 0.30]{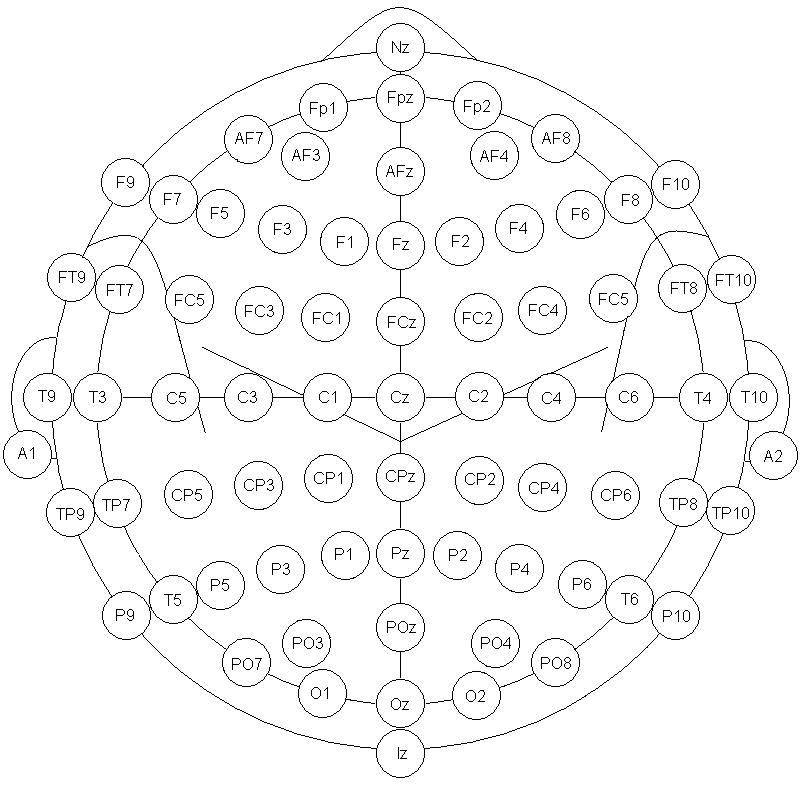}
\caption{Electrode mapping system 10-20 (mindmirror.com).}
\label{fig:eeg1020}
\end{figure}

As mentioned earlier in this study we work in collaboration with the PON Hospital, Jakarta. Using two EEG tools (with a different brand, Xltek and Biologic). EEG signals from 62 persons as the object data are recorded. The proportion of stroke data are 32 and for normal data are 30. The age range of data objects between 29 to 72 years, 37 persons male and 25 persons female. In this study, we also assisted by two neurologists to diagnose the EEG signal and the stroke severity level of the patient’s. For each recording, the task took a 15 minutes duration time in total. For each EEG measurement consist of five events condition. The events include: open eyes without stimulation, closed eyes without stimulation, and closed eyes with a flash of light stimulation in three variants of frequencies (5 Hz, 10 Hz, and 15 Hz). The sampling rate was used in the recording process using Xltek is 512Hz. On the other hand sampling frequency when used Biologic was 512Hz on some of part of data and 256Hz for the rest. Both of tools use 33 electrodes and record 33 channels. For details the 29 channels that has been registered are: FP1, F3, C3,P3, O1,F7,T3,T5, A1, FP2, F4, C4, P4, O2, F8, T4, T6, A2, FZ, CZ, PZ, OZ, FPZ, PG1, PG2, T1, T2, LEOG, REOG, ECG, Event, Reff, and AUX8. Each mapping location for EEG electrodes given by Fig. \ref{fig:eeg1020} \cite{mindmirror}. On this study, we only use 2 channels EEG (C3 and OZ), and 2 electrooculography channels (left EOG and right EOG).

\subsection{Preprocessing}
From the literature, it is known that the ischemic stroke is caused by obstruction. Obstruction makes the CBF level of stroke patients much lower than CBF level of a normal person. Low CBF levels have an impact on slowing the EEG signal of stroke patients. Based on these facts, the focus of the analysis of the EEG signals should be on the slow wave signals. Several other studies on the analysis of frequency bands involves four major waves with band frequencies below than 30Hz (1$-$4Hz delta, 4$-$8Hz theta, 8$-$12Hz alpha, and 12$-$30Hz beta).
The lowest sampling rate when EEG was recorded in our study is 256 Hz. With the sampling rate of 256Hz we can obtain PSD from band frequency between 0Hz to the 128Hz, Of course, this supports the action to distinguish EEG stroke from the EEG control (normal). To make the procedure be fit and much efficient with our goal, we down sample the EEG data and EOG data to the 64Hz. After down sampling from 62 data objects, we have 62x57600 matrix for each channel. On the other hand, the class label represented as 0 for normal and 1 for stroke. Array data to store the label is 62x1.

\subsection{Feature Extraction}
After preprocessing data, we go through to the feature extraction stage. As an input of this stage is 4 channels (C3, Oz, LEOG, and REOG), 62x57600 raw data. As a result from this stages is only 24 feature for each data object. The final data structure for the feature in our study is 62x24. Our 'handcrafted' feature as result from this stage are composed of relative power band frequency, variance, correlation aspect, kurtosis, entropy, spectral mean, and exponent of fractal.

\subsection{Classifier Evaluation}
In this study, we only have 62 data objects. This size is too small if we used k-fold cross validation scenario with a limited number of k, deal with this situation we apply Leave One Out (LOO) cross validation technique to evaluate the performance of each classifier in experiments. The LOO scenario set a scheme: one data as a test and the others data as training upon each round. In this study, each round has 1 data object as a data test and remaining 61 data objects as training data. This condition was repeated 62 rounds so that for each data being once time into test data and 61 times as training data. In our evaluation, this 62 times LOO (full round) was repeated for 5 times. For each full round repetition, each parameter of matrices performance was calculated and then assessed from the average value. Parameters performance that was measured in this experiment is Accuracy (Acc.), Sensitivity (Sens.), Specificity (Spec.), F-Score, Precision (Prec.), and Recall (Rec.).


\section{Experiment Result} 
\begin{figure}
\includegraphics[scale = 0.62]{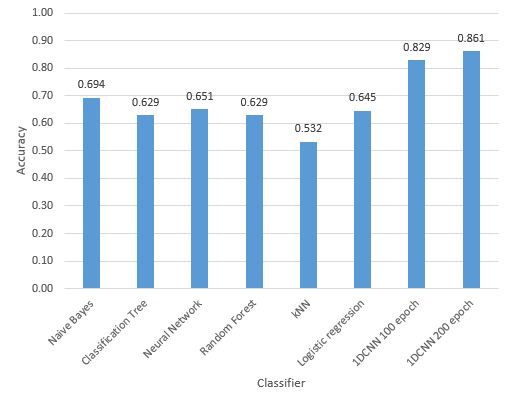}
\caption{Accuracy Level.}
\label{fig:acc}
\end{figure}

Our experiment runs on two operating system environments, Ubuntu 14.04 for implementation 1DCNN and Windows 8 to running Orange for data mining applications. Computer hardware was used in this study have a specification a Quad Core i7 CPU (clocked size 2.4GHz) with 4GB of RAM. The experiment evaluates 1DCNN on two different epochs (100 and 200). Table I shows the performance parameters results for each epoch and all of the comparator classifier. Increasing epoch value accompanied by improvement the performance parameter values of 1DCNN. From the experiment with epoch 100 and epoch 200, we get the average of accuracy are 0.829 and 0.861 respectively. These results get an improvement of accuracy compare to the best achievement of the comparator (only 0.69, Naive Bayes). Descending order by the performance based on the average accuracy are 1DCNN, Naive Bayes, Neural Network, Logistic Regression, Classification Tree, Random Forest, and kNN (Fig. \ref{fig:acc}).
\par
From the experiment simple classifier Naive Bayes success to be the best performance for shallow architecture model. The F-Score of Naive Bayes is only 0.725 on the other hand F-score of 1DCNN is 0.861. From overall evaluation parameters, 1DCNN (both of 100 epoch and 200 epoch) always obtain a better result than all of the compared classifiers. The best value of parameters is achieved by 1DCNN when the maximum epoch is 200. The best value for accuracy, sensitivity, specificity, F-score, precision, and recall has level higher than 0.86.
\par
\begin{figure}
\includegraphics[scale = 0.45]{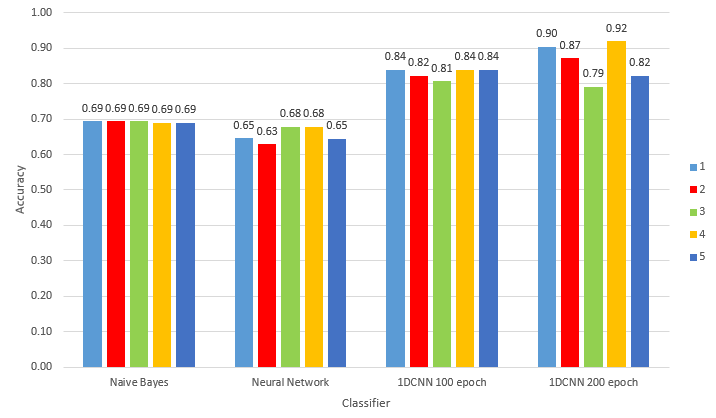}
\caption{Accuracy for each measurment for Naive Bayes. Neural Network and 1DCNN (100 epoch and 200 epoch).}
\label{fig:detepoch}
\end{figure}

 If we analyze further, for each time measurement (Fig. \ref{fig:detepoch} and Table \ref{tbl:epc}) the accuracy level for Naive Bayes is always 0.694. As we all know, construction of Naive Bayes model does not have stochastic aspect, unlike neural architecture models like Neural Network and 1DCNN.  The neural architecture involving random value for the initialization and shuffle object when running the training process (batch training). It makes the classification model for neural architecture possible to have different final value for train weight. Fig. \ref{fig:detepoch} shows accuracy value of the neural network and 1DCNN on five times measurement can be different, nevertheless for 1DCNN the accuracy level always higher than accuracy level of Naive Bayes. The best accuracy from 1DCNN is 0.919. The detail performance from each epoch for Naive Bayes, Neural Network (with 200 epochs), and 1DCNN (100 epochs and 200 epochs) given by Table II. From this result, if we compare between NN and 1DCNN, 1DCNN obtain a better result than NN with a lower number of epoch (100 epoch 1DCNN versus 200 epochs NN).
 \par
Furthermore, if we analyzed based on the difficulty of classification task for each class labels, it can be clearly seen that classifying the normal category is relatively easier than identifying stroke category as shown in Table \ref{tbl:conf}. On the best case result, the measurement of the success level of prediction for the normal (healthy) category is always better than stroke class for Naive Bayes, Neural Network, and 1DCNN. True level for normal category prediction is 0.833 for Naive Bayes, 0.767 for Neural Network, and 100\% (perfect prediction) for 1DCNN. On the other hand, the level of prediction success for stroke category are only 0.563 for Naive Bayes, and 0.594 for Neural Network and 0.844 for 1DCNN. The reason for this situation might be because several stroke patients have early stage level of stroke (0$<$NIHSS$\le$ 4). Hence, it was difficult to distinguish them from the normal class. For future study, we must solve this problem.

\begin{table}[]
\caption{Average performance (5 times measurements) }
\label{tbl:detepoch}
\begin{tabular}{ccccccc}
\cline{1-7}
\multirow{2}{*}{Classifier} & \multicolumn{6}{c}{Parameter}\\ \cline {2-7}
\cline {2-7}
& CA    & Sens  & Spec  & F1    & Prec  & Rec \\ \cline {1-7}
\cline {1-7}
NB&0.694&0.833&0.563&0.725&0.641&0.833\\
CT&0.629&0.600&0.656&0.610&0.621&0.600\\
NN&0.655&0.740&0.575&0.675&0.620&0.740\\
RF&0.629&0.667&0.594&0.635&0.606&0.667\\
kNN&0.532&0.500&0.563&0.509&0.517&0.500\\
Logreg&0.645&0.633&0.656&0.633&0.633&0.633\\
1DCNN 100 ep.&0.829&0.829&0.832&0.829&0.835&0.829\\
1DCNN 200 ep.&0.861&0.861&0.865&0.861&0.870&0.861\\
\cline {1-7}
\end{tabular}
\label{tbl:acc}
\end{table}

\begin{table}[]
\caption{Naive Bayes, Neural Network, and 1DCNN (100 epoch and 200 epoch)(5 times measurements)}
\label{tbl:acc}
\begin{tabular}{@{}ccccccc@{}}
\cline{1-7}
\multirow{2}{*}{Classifier} & \multicolumn{6}{c}{Parameter}\\ \cline {2-7}
\cline {2-7}
& CA    & Sens  & Spec  & F1    & Prec  & Rec \\ \cline {1-7}
\cline {1-7}
NB&0.694&0.833&0.563&0.725&0.641&0.833\\\\

NN (1) &0.645&0.733&0.563&0.667&0.611&0.733\\
NN (2) &0.629&0.700&0.563&0.646&0.600&0.700\\
NN (3) &0.677&0.767&0.594&0.697&0.639&0.767\\
NN (4) &0.677&0.767&0.594&0.697&0.639&0.767\\
NN (5) &0.645&0.733&0.563&0.667&0.611&0.733\\
\\

1DCNN 100(1)&0.839&0.839&0.838&0.839&0.839&0.839\\
1DCNN 100(2)&0.823&0.823&0.825&0.822&0.826&0.823\\
1DCNN 100(3)&0.807&0.806&0.812&0.805&0.819&0.806\\
1DCNN 100(4)&0.839&0.839&0.845&0.838&0.853&0.839\\
1DCNN 100(5)&0.839&0.839&0.838&0.839&0.839&0.839\\\\

1DCNN 200(1)&0.903&0.903&0.905&0.903&0.905&0.903\\
1DCNN 200(2)&0.871&0.871&0.877&0.870&0.886&0.871\\
1DCNN 200(3)&0.790&0.790&0.793&0.790&0.794&0.790\\
1DCNN 200(4)&0.919&0.919&0.924&0.919&0.931&0.919\\
1DCNN 200(5)&0.823&0.823&0.827&0.822&0.832&0.823\\

\cline {1-7}
\end{tabular}
\label{tbl:epc}
\end{table}

\begin{table}[]
\caption{Confusion matrix for NB,NN, and 1DCNN 200 epoch (the best case)}
\label{my-label}
\begin{tabular}{@{}ccccccc@{}}
\cline{1-7}
\multirow{3}{*}{Actual Class} & \multicolumn{6}{c}{Prediction Class}\\ \cline {2-7}
& \multicolumn{2}{c}{NB} & \multicolumn{2}{c}{NN} & \multicolumn{2}{c}{1DCNN ep. 200} \\
\cline {2-7}
& Normal & Stroke & Normal & Stroke & Normal & Stroke \\ \cline {1-7}
Normal	&25	&5	&23	&7	&30	&0  \\
Stroke	&14	&18	&13	&19	&5	&27 \\ 
\cline {1-7}
\end{tabular}
\label{tbl:conf}
\end{table}

\section{Conclusion}
The results of the experiment show that deep learning approach 1DCNN has managed to be the best model to distinguishing task between EEG stroke data to the EEG control data. In this study, we apply early stopping and Batch Normalization techniques to accelerate training process of our classification model. The leave-one-out scenario of 1DCNN obtained average accuracy at 0.86 (F-Score 0.861 and precision 0.870). This achievement was achieved by only 200 epoch and using 24 ‘handcrafted’ feature that focused on four channels 2 EEG channels and 2 EOG channels. If we compared to the best result from shallow classifier (Naive Bayes), 1DCNN get an improvement with about 16.8$\%$. From this result, it was demonstrated that EEG is very potential and possible to used to distinguish a person with stroke to the normal people.


\section*{Acknowledgment}

This work is supported by Collaboration Research Grant
funded by Indonesian Ministry of Research and Higher Education, Contract Number: 1720/UN2.R12/JKP.05.00/2016.


\bibliography{library}

\bibliographystyle{abbrv}

\end{document}